\documentclass[letterpaper, 10 pt, conference]{ieeeconf}
\IEEEoverridecommandlockouts
\usepackage{hyperref}
\usepackage{graphicx}
\usepackage{svg}
\usepackage{amsmath}
\usepackage{amssymb}

\title{\LARGE \bf
Inferring Belief States in Partially-Observable Human-Robot Teams
}

\author{Jack Kolb and Karen M. Feigh%
    \thanks{This work was supported by Amazon Consumer Robotics.}
    \thanks{The authors thank Hae Won Park and Animesh Garg for their insights.}
    \thanks{Jack Kolb and Karen M. Feigh are with the College of Engineering at the Georgia Institute of Technology, North Avenue, Atlanta, GA 30332, USA
        {\tt\small \{kolb, karen.feigh\}@gatech.edu}.
    }
}

\begin{document}

\maketitle
\thispagestyle{empty}
\pagestyle{empty}

\begin{abstract}
We investigate the real-time estimation of human situation awareness using observations from a robot teammate with limited visibility. In human factors and human-autonomy teaming, it is recognized that individuals navigate their environments using an internal mental simulation, or \textit{mental model}. The mental model informs cognitive processes including situation awareness, contextual reasoning, and task planning. In teaming domains, the mental model includes a \textit{team model} of each teammate's beliefs and capabilities, enabling fluent teamwork without the need for explicit communication. However, little work has applied team models to human-robot teaming. In this work we compare the performance of two models, logical predicates and large language models, at estimating user situation awareness over varying visibility conditions. Our results indicate that the methods are largely resilient to low-visibility conditions in our domain, however opportunities exist to improve their overall performance.
\end{abstract}

\section{Introduction}

The human factors community recognizes that when individuals conduct taskwork, they construct an internal simulation termed a \textit{mental model}. The mental model broadly informs the individual's abstract understanding of the environment, such as contextual information, the status of the task at hand, contingencies, and actionable plans \cite{rouse1992role}.

In collaborative tasks, the mental model expands to include a \textit{team model} incorporating a prediction teammate beliefs, characteristics, capabilities, and roles \cite{talone2015evaluation, rouse1992role}. By constructing a team model in addition to their own mental model, individuals apply the information they believe their teammates know – and do not know – in their own plans. This allows individuals to work proactively by anticipating teammate needs without explicit communication \cite{tabrez2020survey, langan2000team}.

Recent research interest has sought to apply \textit{team models} for human-robot and human-autonomy teaming \cite{gervits2020toward} \cite{nikolaidis2012human}. Researchers in the human-robot interaction (HRI) community anticipate that an inferred understanding of human teammate beliefs can support a variety of robot planning, comprehension, and decision-making goals \cite{schuster2011research}. Related work has proposed frameworks for constructing team models \cite{scheutz2017framework} \cite{bolton2022fuzzy}, however few work have evaluated these frameworks in human-robot teaming domains.

Human factors researchers often use \textit{situation awareness} as a proxy for pertinent components of the mental model \cite{endsley1988design, schuster2011research}. Situation awareness is an individual's understanding of the current environment along three \textit{levels}: the raw \textit{world state} (e.g., object locations), \textit{contextual information} (e.g., inferred properties and relationships between objects), and \textit{future projection} (e.g., plans and dynamic models). In practice, situation awareness is typically measured \textit{in-situ} by periodically halting the environment and asking the user questions about the task's current state \cite{endsley1988situation}.

Our objective is to evaluate two models that enable robots to predict human teammates' belief states. We define the \textit{belief state} as measurable information of the teammate's situation awareness. We split the problem into two components -- estimating the user's awareness of world state elements, and estimating how that information is used to answer \textit{Level 1} (world state) and \textit{Level 2} (contextual) situation awareness questions. Each model addresses the second component by reasoning over a scene graph of the user's estimated knowledge of the current task.

Concretely, this paper contributes the following:

\begin{enumerate}
    \item An online human-robot teaming domain and situation awareness dataset featuring partial observability and two belief state prediction models -- a \textit{logical predicates} model and a \textit{large language model} (LLM).
    \item Findings that the belief state prediction models were resilient to low-observability conditions and moderately agreed with human user responses.
\end{enumerate}

Our results indicate that both models were successful at inferring user situation awareness. Moreover, both models maintained their performance under varying visibility conditions. The LLM performed on par with the handcrafted \textit{logical predicates} model, indicating that the LLM can be leveraged to create rules for \textit{logical predicates} models, or encapsulate the process entirely.

A project website with a demo, dataset, and source code is available at \url{https://jackkolb.com/tmm-hri}.

\section{Related Works}

The HRI community has primarily designed mental models with task-dependent factors aligned to Endsley's three levels of situation awareness. For example, in a process-driven task (e.g., manufacturing, housework, spaceflight), the mental model would contain the task procedure (\textit{Level 1}), the context of the environment state with respect to the task procedure (\textit{Level 2}), and what that implies for the immediate future (\textit{Level 3}). Contextual information is derived from the world state, which supports higher-level reasoning about the future of the environment and task. In addition, mental models can draw upon prior experiences to form a stronger model of the risks, failure modes, and practical effects of different actions and task strategies \cite{tabrez2020survey}.

Mental models are also believed to play an important role in teaming  \cite{tabrez2020survey}. In human-human teams, people predict the mental model of their human teammates, the \textit{team model}. If an individual believes their teammate’s mental model is incorrect, or suspects their own mental model is outdated, they adapt their plans or communicate with their teammate accordingly. Related work has found that a strong team model is often important for high-performing teams \cite{langan2000team}.

A \textit{team model} also facilitates theory of mind \cite{chella2018human}. Theory of mind is an agent’s understanding of another agent’s beliefs, such as the other’s knowledge, intentions, emotions, and opinions. Theory of mind is generally studied in the context of education, where instructors aim to shape the belief states of their students, and lends itself to orders of intentionality (\textit{``I believe that you know X”}). Theory of mind is inherently team-centric, and has been studied in several close-proximity human-robot collaboration domains \cite{das2023state2explanation, schrum2022mind}.

Most related work on mental models has focused on humans and human-human teams, however in recent years there has been interest in expanding mental model theory to human-robot teams \cite{rouse1992role, hanna2018impact, gervits2018shared, kennedy2007using}. Work in this space has been challenged by the extraordinary complexity of representing human cognition, which remains an existential cognitive science problem. This has led researchers to use simplified domains and team models. This work uses a domain that more closely represents real-world human-robot teaming applications.

\subsection{Logical Predicates Representation}

The most prevalent representation of mental models in the HRI literature uses \textit{logical predicates}, or sets of known logical truths \cite{edgar2023improving, scheutz2013computational, gervits2018shared, scheutz2017framework}. Each predicate is an observed feature of the environment state taken as an absolute truth, e.g., the location of an object and its properties. Complex functions are handcrafted to derive contextual information or model how the environment will change over time. When predicting the mental model of a teammate, a robot agent can represent the teammate's mental model using the same logical predicates framework, and filter the observational inputs by the workspace region visible to the teammate.

The logical predicates model is attractive for environments with low observational uncertainty. When the human teammate is unlikely to misunderstand their observations, and the robot has full observability, the teammate's mental model is a subset of the robot's mental model. With these assumptions a logical predicates model can capture all relevant environment features and answer a wide range of inference questions.

However, the robot's mental model can be misled when it does not contain all information available to the teammate's mental model, for example, in partially-observable conditions. As the logical predicates model is fact-based and cannot model uncertainty, hard-coding reasoning rules to resolve conflicting information can negatively impact downstream planning and reasoning.

Research opportunities currently exist to consider uncertainty within mental models. Specifically, the literature identifies three areas for error: misperception of events, miscomprehension of perceived events, and imperfect mapping of perceived events to the updated mental model \cite{bolton2022fuzzy}.

In this work we assess whether an \textit{LLM} can match a \textit{logical predicates} model at inference over a scene graph, challenging the need for handcrafted inference rules.

\subsection{Markov Decision Process Representation}

Nikolaidis et al. proposed representing the team model as a partially-observable Markov decision process (POMDP), where the environment is represented as a set of states, environment events, transition probabilities, and state rewards \cite{nikolaidis2012human}. In this representation, the mental model is the predicted probability that the human believes the system is in each state. Higher levels of situation awareness can be estimated by rolling out the Markov process and assuming players and opponents will perform actions that maximize their reward. Additionally, the state rewards and transition probabilities can be tuned to individual teammates to personalize the mental model to the teammate’s experience and characteristics.

The POMDP representation meets the criteria of a mental model in that it models the belief state of an agent. However, it does so by inverting the representation from an object-based belief state to a state-based belief state. In a low-dimensional environment, it is feasible to represent all possible states and their transition probabilities. However, the complex environments sought by HRI domains can overwhelm the capabilities of a POMDP. Few works have applied the POMDP representation to real-world scenarios.

\begin{figure*}
    \centering
    \includegraphics[width=1.0\linewidth]{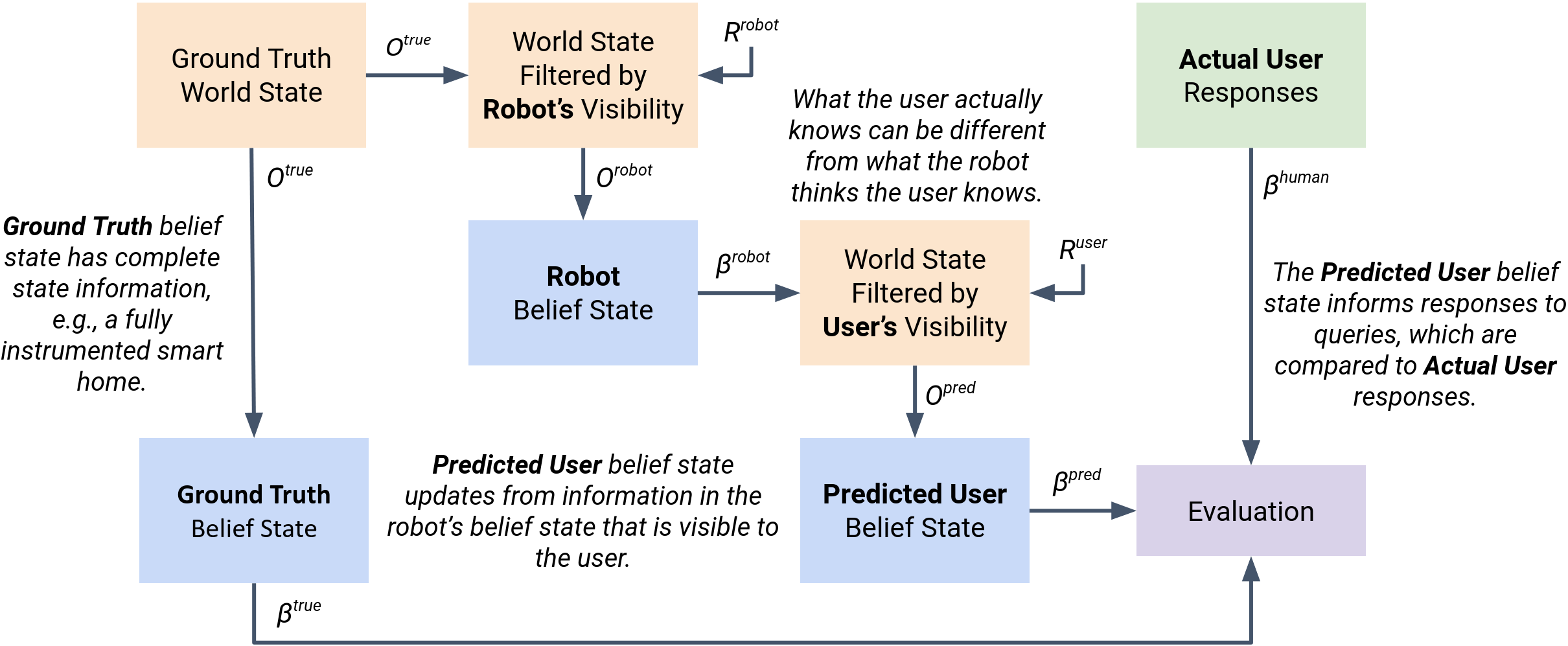}
    \caption{Overview of the predicted user belief state system. The ground truth world state information is filtered by the robot's visibility ($R^{robot})$ and used to update the robot's belief state, $\beta^{robot}$. The robot's belief state is then filtered by the user's visibility to update the predicted user belief state, $\beta^{pred}$. The resulting belief state is what the robot thinks the user is aware of, i.e., a \textit{theory of mind} that can inform downstream reasoning tasks.}
    \label{system-fig}
\end{figure*}

\subsection{Scene Graphs}

In practice, a belief state is built upon a representation of the world state. Prior work typically uses a semantic scene graph of key environment objects and properties, reducing the complexity of downstream reasoning and inference \cite{gervits2020toward}.

A scene graph is defined as a graph $G = \{V, E\}$. Graph nodes $V$ represent environment objects, with properties such as the object's location, class, and instance-level attributes. Graph edges $E$ represent relationships between nodes, with properties such as the semantic relationship between objects (e.g., \textit{onTopOf}). Nodes are not limited to environment objects and can include higher-level concepts such as \textit{rooms} and \textit{structures} \cite{rosinol20203d}. Similarly, edges can include information about what tasks an object is useful for and their decomposition. As current representations of mental models are task-dependent, the scene graph offers a flexible structure for representing physical \textit{and} abstract concepts in the scene.

In a dynamic scene graph (DSG), the scene changes with time and $G$ is updated as the agent receives new information from cameras or other sensors. In partially-observable domains -- i.e., a mobile robot with a camera -- maintaining a DSG is highly challenging because of object impermanence. A DSG must not only track where objects are, but resolve new, moved, and removed object instances. Consequently, no libraries exist for constructing DSGs in partially-observable environments. Instead, related works often extract the scene graph from the simulator itself or simplify the domain to avoid instance-level reasoning (e.g., no two objects of the same class and color). In this work we introduce a basic -- but functional -- model for maintaining a DSG.

\section{Problem Formulation}

Our objective is to map a set of observable environment states $O$ to a set of belief states $\{\beta^{true}, \beta^{robot}, \beta^{pred}\}$ over a time series. The belief states represent task-relevant world state and contextual information about the environment for an oracle ($\beta^{true}$), a robot ($\beta^{robot}$), and a human user ($\beta^{pred}$). $\beta^{true}$ is the ground truth of the environment, and is updated from the full environment state $O_t$ at each timestep $t$. $\beta^{robot}$ uses the subset of $O_t$ that is observable by the robot within an observable region, $R^{robot}$. Lastly, $\beta^{pred}$ is a predicted belief state of the human user from the robot's perspective, updated using $\beta^{robot}$ and the user's observable region $R^{user}$.

We approach the problem by defining two functions, $\Phi$ and $\mathcal{B}$, to update the belief state from observations.

\begin{equation}
    \Phi : O_t, R \mapsto O'_t
\end{equation}

$\Phi$ filters environment observations $O_t$ by a human or robot agent's observable region $R$ to obtain the subset of observations $O'_t$ that are directly observable by the agent. In a 2D domain, if $R$ is a circle of radius $r$, then $O'_t$ is the set of observations from $O_t$ that are within $r$ from the agent. In a 3D domain, $R$ can use the agent's pose and perception capabilities to model the agent's visible region. When inferring belief states, such as $\beta^{pred}$, $O_t$ is extracted from $\beta^{robot}$ and can differ from the true world state.

\begin{equation}
    \mathcal{B} : O'_{t}, \beta_{t-1} \mapsto \beta_{t}
\end{equation}

$\mathcal{B}$ updates a belief state from observations. $\beta$ is the current belief state, and $O'_t$ is the subset of environment observations that are used to update the belief state at a time $t$.

The core research objective of inferring a human teammate's belief state is to identify the function $\mathcal{B}$ that maximizes the agreement between $\beta_{t}$ and the teammate's internal belief state. In this work, we model a human user's belief state ($\beta^{pred}_t$) and score the agreement between $\beta^{pred}_t$ and queries asked of study participants ($\beta^{human}$). In Fig. \ref{system-fig}, $\beta^{true}$ is the ground truth of the environment, equivalent to a robot with full observability.

We evaluate the following hypotheses:

\begin{enumerate}
    \item[\textbf{H1}] A fully-observable logical predicates baseline ($\mathcal{B}^{LP}$) will result in responses to situation awareness queries that agree with user responses.

    \item[\textbf{H2}] The performance of the logical predicates baseline ($\mathcal{B}^{LP}$) at inferring user situation awareness will sharply degrade in partially-observable conditions.

    \item[\textbf{H3}] A large language model ($\mathcal{B}^{LP+LLM}$) will match the logical predicates baseline ($\mathcal{B}^{LP}$) across fully-observable and partially-observable environments.
\end{enumerate}

\section{Methods}

Our study extends the \textit{Overcooked-AI} environment to predict user belief states in partial observability conditions \cite{carroll2019utility}. The environment simulates a restaurant kitchen where several agents collaborate to cook and deliver soups using ingredients scattered across the kitchen's counters. While the domain is a discretized representation of a cooking task, it presents a team-oriented structure and a fast-paced, engaging task. Additionally, the discrete nature is ideal for evaluating the \textit{logical predicates} model, which the HRI community has only explored in domains that are even more simplistic \cite{gervits2020toward}.

In \textit{Overcooked-AI}, agents in a human-robot team perform a sequence of subtasks to cook soups. Agents first move three ingredients (onions or tomatoes) from countertops to a pot. The pot then cooks for $10$ seconds, and once complete, an agent `plates' the soup using a dish. Lastly, the plated soup is brought to a designated serving station, where it leaves the gameboard. The process repeats until all ingredients have been used or $90$ seconds have elapsed. Users are told to focus on completing all possible soups. While each user's time to complete and quantity of completed soups are available in the dataset, we only used user performance to screen out participants who ignored the task. Participants were filtered out if they answered no situation awareness questions correctly, or failed to cook at least one soup in any of the four kitchen layouts.

The agents share a workspace, but cannot communicate, meaning the team benefits from agents being proactive in their work and considerate of their teammate's belief state. While related work has used reinforcement learning for the robot agent \cite{carroll2019utility}, our preliminary work found that a task-oriented state machine with A* planning was effective across our kitchen layouts. In this work, we are focused on predicting the user belief state ($\beta^{pred}$), not applying that prediction to planning and decision-making.

\subsection{Partial Observability}

We adapted the \textit{Overcooked-AI} environment to simulate real-world conditions by introducing partial observability. Our visibility function, $\Phi$, limits an agent's field of view to a region defined by an agent-specific parameter $R$. The parameter includes a type (\textit{V}, \textit{O}, or \textit{D}) and a radius:

\begin{itemize}
    \item \textit{V}-type is a $90^{\circ}$ conic view facing the agent's direction, which represents a narrow field of view.
    \item \textit{O}-type is a $360^{\circ}$ circular view of the surrounding area, and can model full observability when the radius exceeds the environment dimensions.
    \item \textit{D}-type is a $180^{\circ}$ semi-circular view of the area in front of the agents, and can model half observability when the radius exceeds the environment dimensions.
\end{itemize}

\begin{figure}
    \centering
    \includegraphics[width=1.0\linewidth]{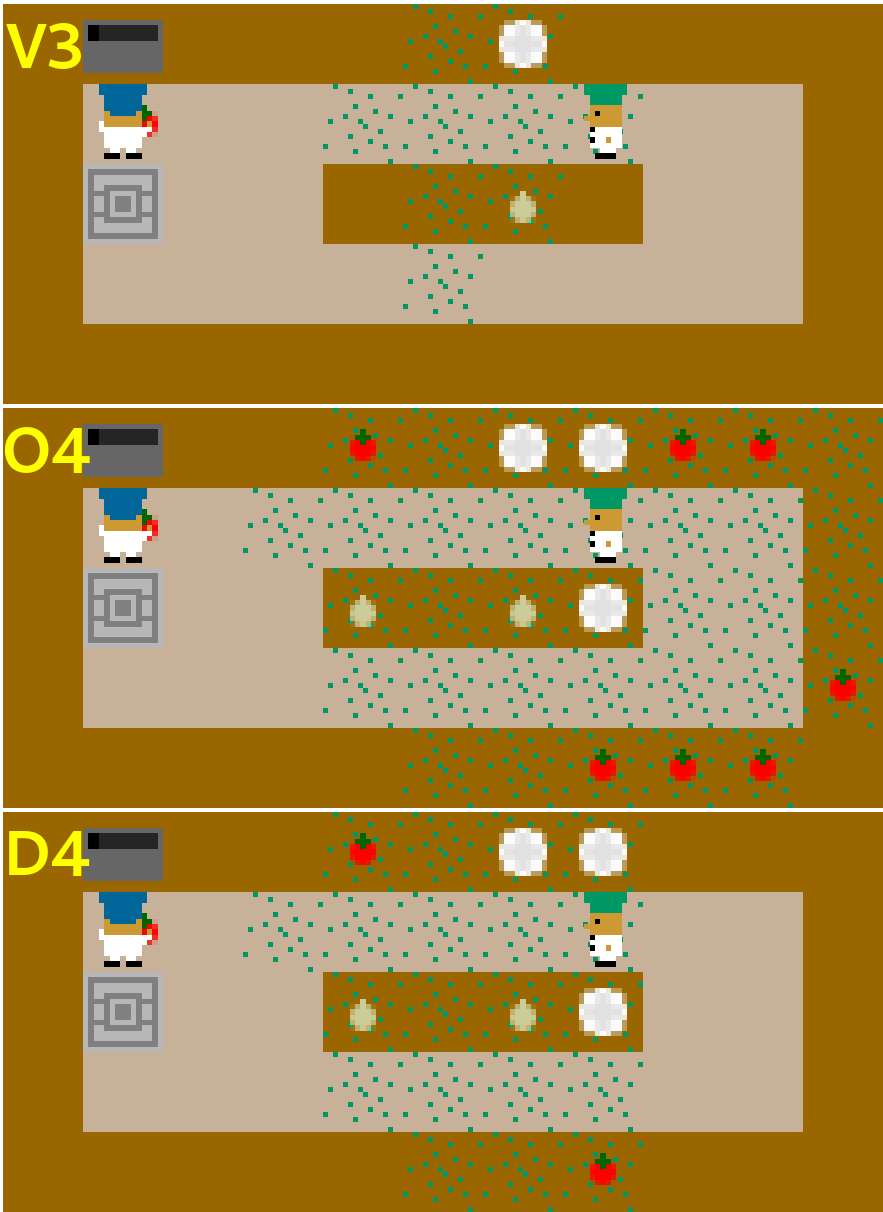}
    \caption{The task domain as shown to the user (green agent), with three visibility types and radii (e.g., \textbf{V3} indicates V-type with radius $3$). The green shaded region is visible to the user. Each type is a unique shape representing a common field of view, and the radius indicates the extent of the visible region. For our experiment, all users had \textbf{D4} visibility, and we varied the visibility of the robot (blue agent) in our post-hoc analysis.}
    \label{visibility-fig}
\end{figure}

Fig. \ref{visibility-fig} shows an example of each visibility type. Agents are able to see the environment objects within their visible region $R$. As we would expect real-world agents to track teammates via sound \textit{and} sight, as well as have a strong knowledge of the static kitchen components, agents can always see the locations of their teammate, the kitchen floorplan, and the kitchen appliances (pot, serving station).

\begin{figure*}
    \centering
    \includegraphics[width=1.0\linewidth]{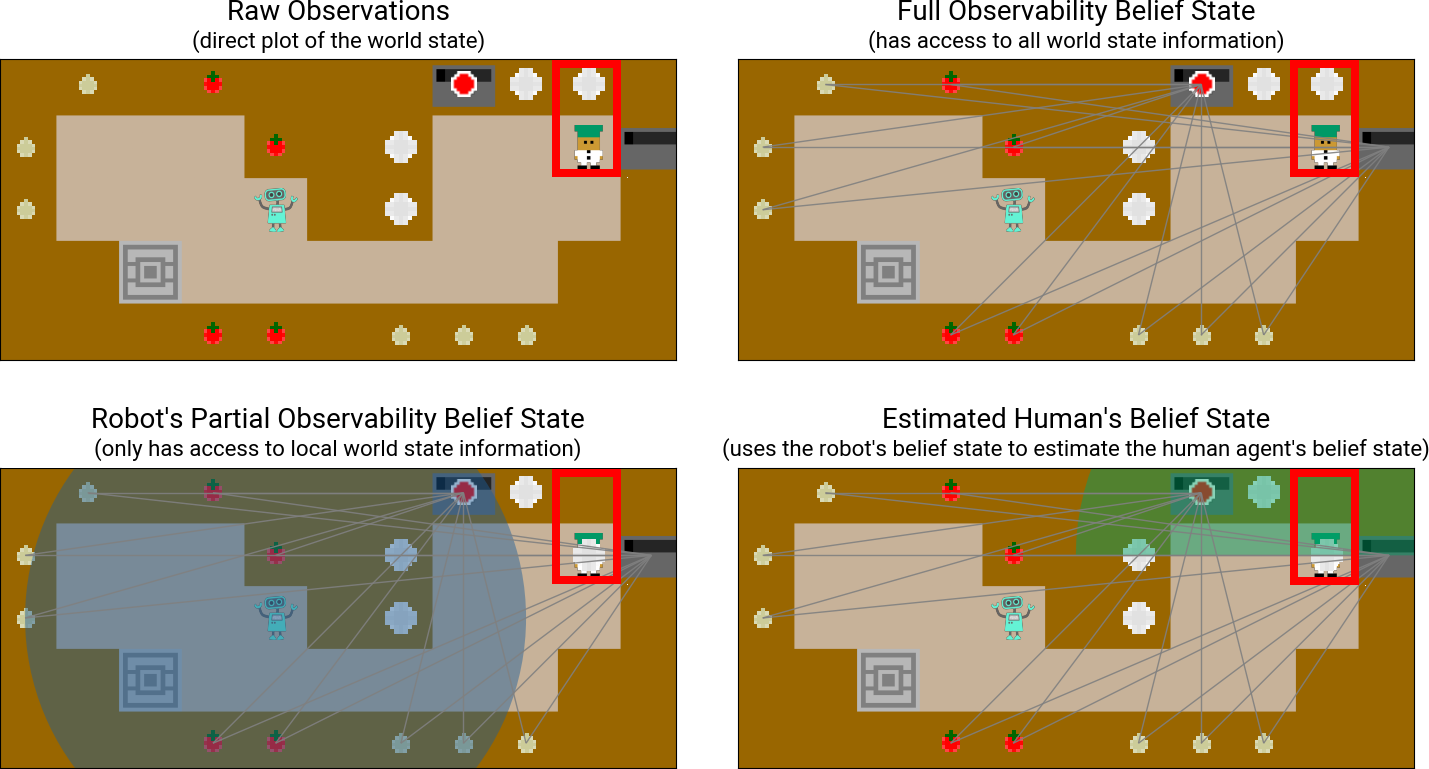}
    \caption{Four views of the same scene. The \textit{Raw Observations} view shows the direct world state $O_t$, the \textit{Full Observability Belief State} ($\beta^{true}$) shows the ground truth beliefs, the \textit{Robot's Partial Observability Belief State} ($\beta^{robot}$) is updated from the world state within the blue region around the robot ($R^{robot}$), and the \textit{Estimated Human's Belief State} ($\beta^{user}$) is updated from the robot's belief state filtered by the green region around the user ($R^{user}$). Grey edges indicate that two objects can be used with each other at this time. A false belief is highlighted in red: the user set a plate upon a counter, however since it occurred outside the view of the robot, the robot's belief state and the estimated human belief state still show the plate as being held.}
    \label{graph-fig}
\end{figure*}

\subsection{Object Permanence in the Scene Graph}

We considered belief states as two components -- a scene graph constructed from the filtered environment observations $O'_t$, and an inference over the scene graph to infer contextual information. However, observations do not inherently track objects, and without object permanence the agent's belief state cannot support objects moving in and out of the visible region. Thus, we require a function to track objects across updates to the scene graph and resolve conflicts. Object permanence is still a challenging open research problem \cite{van2023tracking}.

In addition to object permanence, the cooking environment is challenging because of \textit{object transformation}. When ingredients are used to make a soup, the ingredients simply no longer exist. Similarly, soups leave the gameboard when they are placed on the serving station. Our system must not only track objects through time, but also identify when objects have transformed and predict which objects were transformed. To help simplify this problem, we assume that all ingredient objects are known at the task's initialization.

Object permanence was maintained using a structured three-pass logical approach. In the first pass, static environment objects are identified and matched to their corresponding known object. While this naive case can be tricked by swapping an item with an identical item, it is not necessary for our belief states to identify that edge condition.

The second pass matches objects to their closest known object. Similar to the naive case, the pass matches each remaining observed object to the closest remembered object of the same class. This is most effective with frequent low-ambiguity observations, which is feasible for a two-agent team. To resolve objects that were picked up by agents, the pass defaults to the closest unmatched known object.

The final pass addresses transformed objects. These are observed objects that have no fit to a known object (i.e., a newly-made soup), or known objects that have no observed match (i.e., a plated soup that was delivered). Due to the partially-observable environment, it can be challenging to determine which objects transformed into other objects, however many cases can be inferred by tracking a soup's ingredients. For example, a new ``tomato, onion, and tomato soup'' object implies that two known tomatoes and one known onion no longer exist on the gameboard. As each agent's belief state is initialized with the true gameboard state, and no ingredients are added during gameplay, this pass can reasonably identify which ingredients were used to create a soup and when a soup has likely been served.

Due to the freedom users have to select ingredients and complete the task, no object permanence system can be perfect in all partially-observable environments.

Fig. \ref{graph-fig} shows a reconstruction of the gameboard using three belief states. Notably, the figure shows that the robot's belief state $\beta^{robot}$ mistracks a plate that the user set down outside of the robot's visible region $R$, which cascades to a misbelief in the predicted user belief state $\beta^{pred}$.

\subsection{Inferring Context and Situation Awareness}

The value of modeling a teammate's belief state lies in applying the belief state to correctly infer additional contextual information and answer questions. We use the SAGAT method \cite{endsley1988design} to obtain the user's real belief state, $\beta^{human}$, by periodically pausing the game and asking the user two questions about the environment every $30$ seconds. We include questions about the first two levels of situation awareness -- world state understanding and contextual information. All situation awareness questions were multiple choice and randomly chosen from a set question bank.

As shown in Fig. \ref{system-fig}, the user responses are compared to the predicted belief state $\beta^{pred}$ to evaluate the performance of the belief state function $\mathcal{B}$. We compare two functions, $\mathcal{B}^{LP}$ and $\mathcal{B}^{LP+LLM}$.

For the logical predicates model, $\mathcal{B}^{LP}$, handcrafted rules use the current scene graph to derive responses to each situation awareness question. For example, to answer the contextual information question \textit{``How many more soups can be made/delivered?''}, $\mathcal{B}^{LP}$ calculates the maximum number of remaining soups using the ingredients and plates known to $\beta^{pred}$. Consequently, the \textit{logical predicates} model is unable to answer questions that are not handcrafted beforehand. The model's specific rules are detailed on the project webpage.

$\mathcal{B}^{LP+LLM}$ uses $\mathcal{B}^{LP}$ to construct the predicted scene graph component of $\beta^{pred}$, and an LLM (GPT4 \cite{achiam2023gpt}) to reason upon $\beta^{pred}$ to answer situation awareness questions. The model is prompted with a description of the game and the scene graph $\beta^{pred}$, as well as the query \textit{``You are A1 [...] Please answer the question using only one of responses below [...] What is your answer?''} To fairly compare $\mathcal{B}^{LP}$ and $\mathcal{B}^{LP+LLM}$, the models do not consider prior world states nor data from other participants, although exploring alternative methods of constructing the inferred scene graph is an opportunity for future work. The primary benefit of using an LLM for this task is the LLM's ability to answer open vocabulary questions about the scene, which removes the human effort needed to define the rules of an LP model (albeit at a high computational cost). Additionally, an LLM can distill its reasoning into a logical format compatible with a $\mathcal{B}^{LP}$ model for efficient repeat queries. Our specific LLM prompt is on the project webpage.

\subsection{Scoring and Evaluation}

Belief states are scored by comparing the model's responses of the belief states to situation awareness questions. Correct responses are awarded one point, and incorrect responses zero. For questions based on spatial reasoning, near-correct answers received 0.5 points. For example, for a question asking \textit{``Where is the closest tomato?''}, responding \textit{``Center''} when the correct answer was one tile off-center scored $0.5$. Partial-credit was used because the situation awareness questions were multiple choice, so a distance metric was infeasible. The scoring function is defined as:

\begin{equation}
    Score_{u,l} = \frac{\sum\limits_{q,t \in N} score_q(\beta^{A}_t, \beta^{B}_t)}{|N|}
\end{equation}

Where $q$ is a situation awareness question asked to the user $u$ on layout $l$ at time $t$, $N$ is the total number of situation awareness questions asked, and $score_q (\beta^{A}_t, \beta^{B}_t) \mapsto \mathbb{R}$ is the scoring function for $q$ using two belief states $\beta^{A}_t$ and $\beta^{B}_t$. Our scoring function was symmetric, such that $score_q(\beta^{A}_t, \beta^{B}_t) = score_q(\beta^{B}_t, \beta^{A}_t)$.

\section{User Study Design}

We evaluated our methods with data collected through an online user study approved by Georgia Tech's IRB. Participants were recruited from the research platform Prolific \cite{palan2018prolific}. The study's screening criteria included accessing the study using a personal computer, being within the ages of 18-89, self-identification as \textit{not} part of any vulnerable group, being physically located in the United States, and having English as a first language. The study took approximately 20 minutes to complete, and participants were compensated $\$5$ USD.

\begin{figure}[t!]
    \centering
    \includegraphics[width=1.0\linewidth]{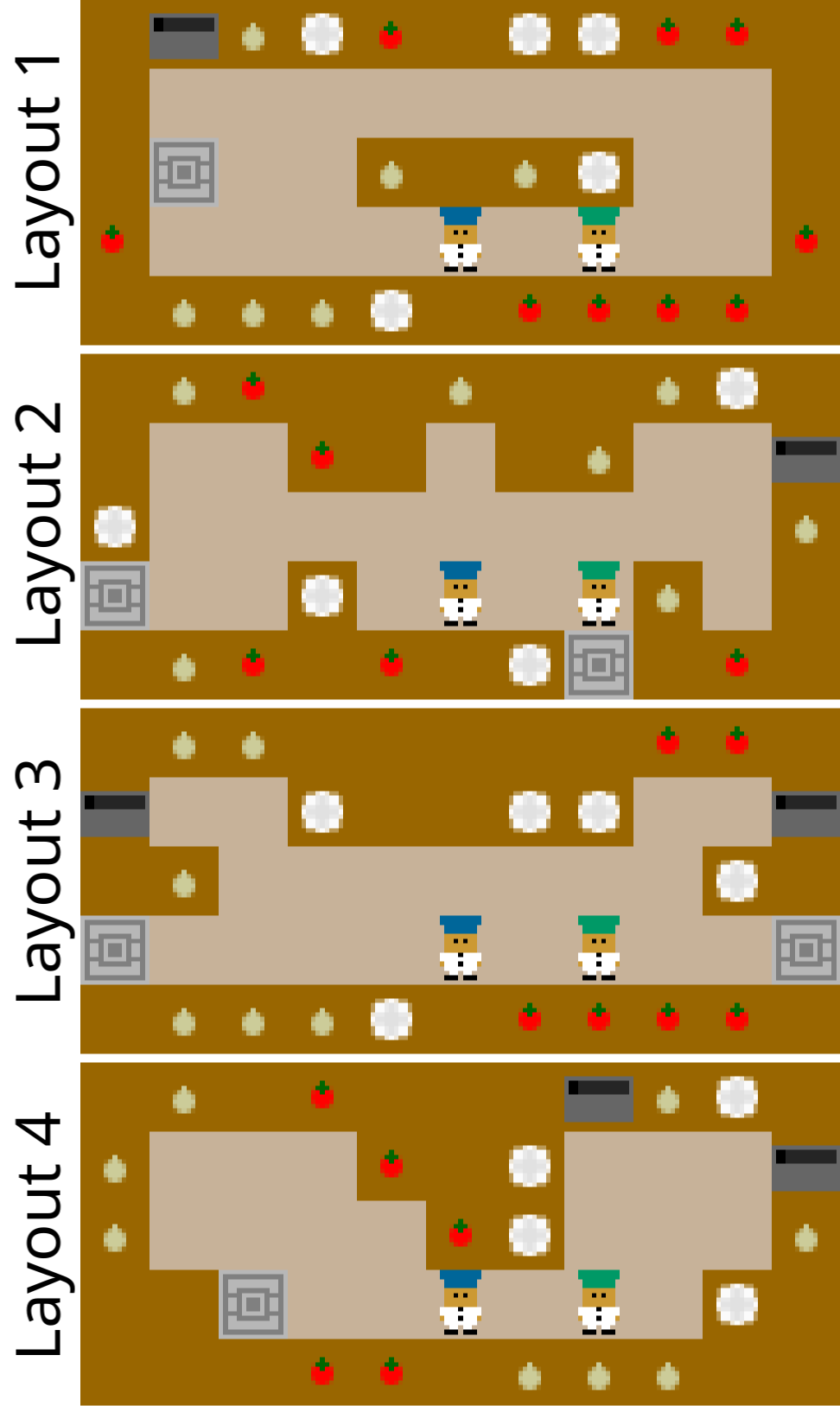}
    \caption{Starting gameboard of each layout in the user study.}
    \label{layout-fig}
\end{figure}

Participants completed two practice runs and then four trials with our experiment domain. Each trial used a different kitchen layout shown in Fig. \ref{layout-fig}. The game's state was recorded at each frame ($10$hz), as well as responses to situation awareness questions.

Post-hoc, the game state histories were replayed and the \textit{logical predicates} model was used to construct $\beta^{true}$, $\beta^{robot}$, and $\beta^{pred}$ across a range of visibilities for $R^{robot}$. The $\mathcal{B}^{LP}$ and $\mathcal{B}^{LP+LLM}$ models were then used to compare responses derived from the models to the actual user responses.

\begin{figure}
    \centering
    \includegraphics[width=1.0\linewidth]{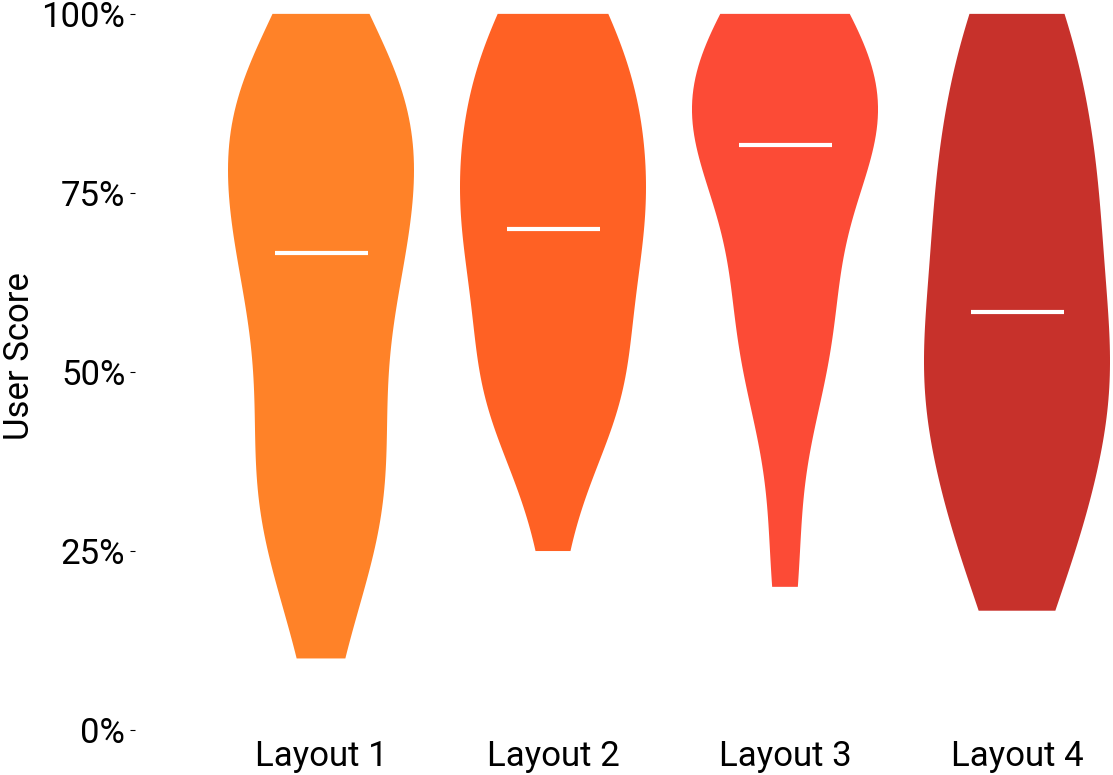}
    \caption{The score distributions of user responses to situation awareness questions ($\beta^{human}$) with respect to the ground truth ($\beta^{true}$). The broad variances show the task is appropriately difficult. Fig. \ref{layout-fig} shows the layouts.}
    \label{user-performance-fig}
\end{figure}

\section{Results \& Discussion}

We continued recruiting participants until $30$ participants passed the minimum participation conditions. A total of $42$ participants were recruited ($50\%$ female, ages $22-74$), of whom $30$ are considered in the dataset and model evaluation.

\subsection{User Performance}

We first judge whether the task was appropriately difficult. Fig. \ref{user-performance-fig} shows the distribution of user scores with respect to the ground truth across the four evaluated kitchen layouts. While \textit{Layout 3} appears to be easier than the other layouts, the other score distributions have similar medians, indicating consistent difficulty. Despite several perfect scores at each layout, the broad score distributions indicate that the task was neither trivial nor overwhelming for users.

Notably, very few users scored perfectly (scores of $100\%$), and no users scored consistently well across all layouts. This suggests that users were generally unable to fully track the environment, motivating the potential benefits of predicting user belief states in this partially-observable domain.

\subsection{Fully-Observable Robot Belief State}

Hypothesis \textbf{H1} evaluates how accurately $\mathcal{B}^{LP}$ with full observability can predict the situation awareness responses of a human with partial observability. Unlike our evaluation of the \textit{User Performance}, wherein the human user was scored against the ground truth oracle, in this case the robot does not assume that the user has any knowledge of the environment outside of what the robot knows the user has explicitly seen.

We find that the scores from the predicted user belief state ($\beta^{pred}$) with respect to the actual user responses ($\beta^{human})$ are almost identical to Fig. \ref{user-performance-fig}. This indicates that there were few instances in our study where a model assuming perfect user perception could have had a false belief, i.e., environment changes that would have impacted the situation awareness responses were quickly observed by users, although seemingly not always recalled.

Similarly, we find that the full observability model's performance averaged between $55\%$ and $75\%$. This suggests that users had nuanced perception, comprehension, or memory that was not captured by $\mathcal{B}^{LP}$. Opportunities exist for alternative models to utilize human factors that can contribute to a user's belief state.

Thus, we affirm our hypothesis \textbf{H1}.

\begin{figure}[b!]
    \centering
    \includegraphics[width=1.0\linewidth]{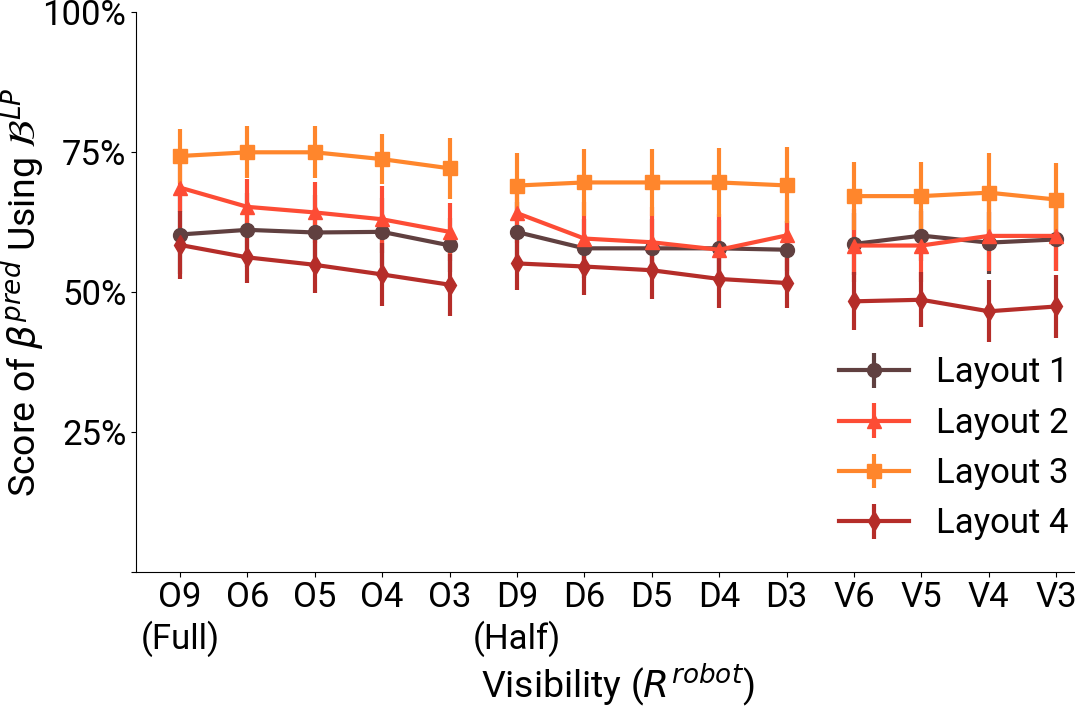}
    \caption{Line plot of the performance of $\beta^{pred}$ using $\mathcal{B}^{LP}$ with respect to the user responses, across visibility parameters $R^{robot}$. Error bars indicate variance. While performance declines between the three visibility types, there is not a significant decline between visibility conditions.}
    \label{visibility-lp-fig}
\end{figure}

\subsection{Partial Robot Observability}

Hypothesis \textbf{H2} addresses how $\mathcal{B}^{LP}$ affects the performance of $\beta^{robot}$ across a range of visibilities. For a belief state function $\mathcal{B}$ to be feasible in human-robot teaming domains, it is valuable to see how the function's performance declines with decreasing visibility.

Fig. \ref{visibility-lp-fig} shows the performance of $\mathcal{B}^{LP}$ across $14$ visibility conditions. Counter to our hypothesis \textbf{H2}, the performance of $\mathcal{B}^{LP}$ did not significantly decrease between high visibility and low visibility conditions (2-sample t-test, p$>$$0.1$ for all). We believe this due to our logical predicates implementation accurately resolving conflicting beliefs during the task. The findings support using $\mathcal{B}^{LP}$ as a baseline for more complex partially-observable domains.

Thus, we refute our hypothesis \textbf{H2}.

\begin{figure}
    \centering
    \includegraphics[width=1.0\linewidth]{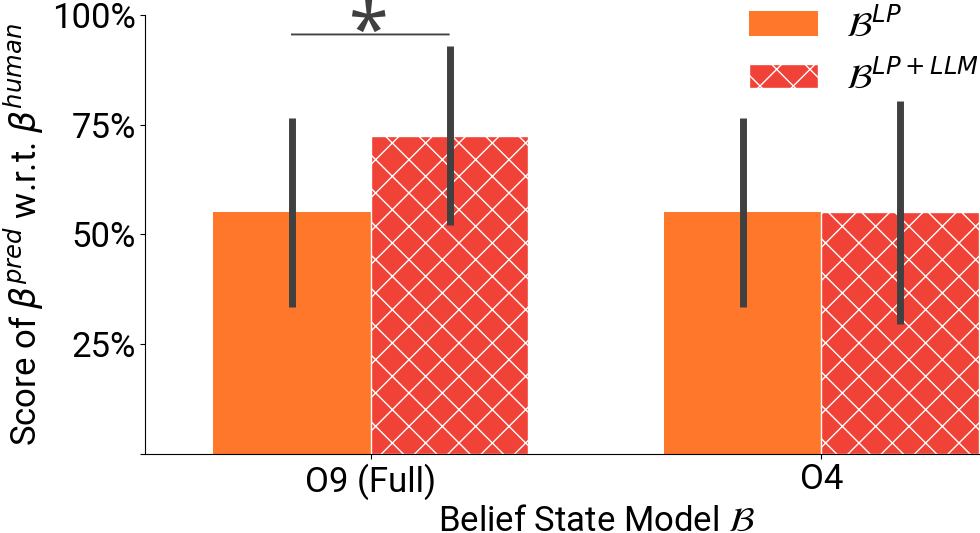}
    \caption{Mean performances of $\beta^{pred}$ constructed using $\mathcal{B}^{LP}$ and $\mathcal{B}^{LP+LLM}$ over two observability conditions. Higher scores indicate superior performance, error bars show variance. $\mathcal{B}^{LP+LLM}$ matched or exceeded $\mathcal{B}^{LP}$ when scored against user responses, indicating that $\mathcal{B}^{LP+LLM}$ is able to infer an ruleset similar to $\mathcal{B}^{LP}$'s handcrafted rules.}
    \label{llm-fig}
\end{figure}

\subsection{Large Language Model}

Hypothesis \textbf{H3} explores whether an LLM can predict and reason about the user's belief state as a zero-shot system. The open vocabulary reasoning of LLMs departs from the handcrafted rules of \textit{logical predicates} models, adding value for flexible real-world domains. As a first step, we are interested in whether an LLM can match the performance of $\mathcal{B}^{LP}$.

Fig. \ref{llm-fig} shows the performance of $\mathcal{B}^{LP}$ and $\mathcal{B}^{LP+LLM}$ at predicting user responses to situation awareness questions. The models were presented with the same scene graph and prompted with which answer the user would choose. Surprisingly, $\mathcal{B}^{LP+LLM}$ outperformed $\mathcal{B}^{LP}$ in full observability (2-sample t-test, p=$0.09$), while the models had near-identical performances given partial-visibility (O4). This suggests that LLMs are capable of inferring similar rules to handcrafted models in zero-shot. The findings support using LLMs to predict user belief states, and as a baseline in future work.

Thus, we affirm our hypothesis \textbf{H3}.

\section{Conclusion}

For decades, the \textit{logical predicates} model has been the de-facto representation of world states, and has recently been extended to represent belief state prediction for human-robot teaming. The logical structure makes the model compatible with state machines, linear temporal logic planners, and case-based reasoning, which is attractive for real-time systems seeking transparent knowledge representations.

However, the \textit{logical predicates} model does not consider the probabilistic and uncertain nature of human cognition. As such, it struggles to represent individual human differences in their interpretations of the environment. Moreover, very few works have operationalized the model to human-robot teaming domains, and the literature typically assumes that robot agents have full observability in their environments. As human-robot teams engage in complex environments, misinterpretation can be consequential to the team's tasks.

In this work we explored using a \textit{logical predicates} belief state model in human-robot teaming. We defined the problem of predicting a human teammate's belief state using the model's responses to situation awareness questions. Through a user study we collected a dataset of responses at various points throughout a collaborative cooking game, and evaluated a \textit{logical predicates} model and an LLM at predicting user responses under several degrees of visibility.

Key takeaways from this work include:

\begin{enumerate}
    \item The \textit{logical predicates} model resulted in a scene graph where neither inference model scored above $50-75\%$, motivating the integration of human factors techniques to construct predicted scene graphs for teammates.

    \item The performance of the \textit{logical predicates} model did not significantly degrade between fully-observable and partially-observable conditions, supporting the model's use in more realistic partially-observable domains if a careful implementation is feasible.

    \item The performance of the \textit{logical predicates} model was matched by a \textit{large language model} in a zero-shot manner, suggesting that LLMs can be leveraged to construct or replace \textit{logical predicate} models.
\end{enumerate}

\bibliographystyle{plain}
\bibliography{paper.bib}

\begin{thebibliography}{10}

\bibitem{achiam2023gpt}
Josh Achiam, Steven Adler, Sandhini Agarwal, Lama Ahmad, Ilge Akkaya, Florencia~Leoni Aleman, Diogo Almeida, Janko Altenschmidt, Sam Altman, Shyamal Anadkat, et~al.
\newblock Gpt-4 technical report.
\newblock {\em arXiv preprint arXiv:2303.08774}, 2023.

\bibitem{bolton2022fuzzy}
Matthew~L Bolton, Elliot Biltekoff, and Kevin Byrne.
\newblock Fuzzy mental model finite state machines: A mental modeling formalism for assessing mode confusion and human-machine “trust”.
\newblock In {\em 2022 IEEE 3rd International Conference on Human-Machine Systems (ICHMS)}, pages 1--4. IEEE, 2022.

\bibitem{carroll2019utility}
Micah Carroll, Rohin Shah, Mark~K Ho, Tom Griffiths, Sanjit Seshia, Pieter Abbeel, and Anca Dragan.
\newblock On the utility of learning about humans for human-ai coordination.
\newblock {\em Advances in neural information processing systems}, 32, 2019.

\bibitem{chella2018human}
Antonio Chella, Francesco Lanza, Arianna Pipitone, Valeria Seidita, et~al.
\newblock Human-robot teaming: Perspective on analysis and implementation issues.
\newblock In {\em AIRO@ AI* IA}, pages 12--17, 2018.

\bibitem{das2023state2explanation}
Devleena Das, Sonia Chernova, and Been Kim.
\newblock State2explanation: Concept-based explanations to benefit agent learning and user understanding.
\newblock {\em Advances in Neural Information Processing Systems}, 36:67156--67182, 2023.

\bibitem{edgar2023improving}
Gwendolyn Edgar, Matthew McWilliams, and Matthias Scheutz.
\newblock Improving human-robot team performance with proactivity and shared mental models.
\newblock In {\em Proceedings of the 2023 International Conference on Autonomous Agents and Multiagent Systems}, pages 2322--2324, 2023.

\bibitem{endsley1988design}
Mica~R Endsley.
\newblock Design and evaluation for situation awareness enhancement.
\newblock In {\em Proceedings of the Human Factors Society annual meeting}, volume~32, pages 97--101. Sage Publications Sage CA: Los Angeles, CA, 1988.

\bibitem{endsley1988situation}
Mica~R Endsley.
\newblock Situation awareness global assessment technique (sagat).
\newblock In {\em Proceedings of the IEEE 1988 national aerospace and electronics conference}, pages 789--795. IEEE, 1988.

\bibitem{gervits2018shared}
Felix Gervits, Terry~W Fong, and Matthias Scheutz.
\newblock Shared mental models to support distributed human-robot teaming in space.
\newblock In {\em 2018 aiaa space and astronautics forum and exposition}, page 5340, 2018.

\bibitem{gervits2020toward}
Felix Gervits, Dean Thurston, Ravenna Thielstrom, Terry Fong, Quinn Pham, and Matthias Scheutz.
\newblock Toward genuine robot teammates: Improving human-robot team performance using robot shared mental models.
\newblock In {\em Aamas}, pages 429--437, 2020.

\bibitem{hanna2018impact}
Nader Hanna and Deborah Richards.
\newblock The impact of multimodal communication on a shared mental model, trust, and commitment in human--intelligent virtual agent teams.
\newblock {\em Multimodal Technologies and Interaction}, 2(3):48, 2018.

\bibitem{kennedy2007using}
William~G Kennedy and J~Gregory Trafton.
\newblock Using simulations to model shared mental models.
\newblock In {\em Proceedings of the eighth international conference on cognitive modeling}, pages 253--254. Taylor \& Francis Psychology Press Ann Arbor, 2007.

\bibitem{langan2000team}
Janice Langan-Fox, Sharon Code, and Kim Langfield-Smith.
\newblock Team mental models: Techniques, methods, and analytic approaches.
\newblock {\em Human factors}, 42(2):242--271, 2000.

\bibitem{nikolaidis2012human}
Stefanos Nikolaidis and Julie Shah.
\newblock Human-robot teaming using shared mental models.
\newblock {\em ACM/IEEE HRI}, 2012.

\bibitem{palan2018prolific}
Stefan Palan and Christian Schitter.
\newblock Prolific. ac—a subject pool for online experiments.
\newblock {\em Journal of Behavioral and Experimental Finance}, 17:22--27, 2018.

\bibitem{rosinol20203d}
Antoni Rosinol, Arjun Gupta, Marcus Abate, Jingnan Shi, and Luca Carlone.
\newblock 3d dynamic scene graphs: Actionable spatial perception with places, objects, and humans.
\newblock {\em arXiv preprint arXiv:2002.06289}, 2020.

\bibitem{rouse1992role}
William~B Rouse, Janis~A Cannon-Bowers, and Eduardo Salas.
\newblock The role of mental models in team performance in complex systems.
\newblock {\em IEEE transactions on systems, man, and cybernetics}, 22(6):1296--1308, 1992.

\bibitem{scheutz2013computational}
Matthias Scheutz.
\newblock Computational mechanisms for mental models in human-robot interaction.
\newblock In {\em Virtual Augmented and Mixed Reality. Designing and Developing Augmented and Virtual Environments: 5th International Conference, VAMR 2013, Held as Part of HCI International 2013, Las Vegas, NV, USA, July 21-26, 2013, Proceedings, Part I 5}, pages 304--312. Springer, 2013.

\bibitem{scheutz2017framework}
Matthias Scheutz, Scott~A DeLoach, and Julie~A Adams.
\newblock A framework for developing and using shared mental models in human-agent teams.
\newblock {\em Journal of Cognitive Engineering and Decision Making}, 11(3):203--224, 2017.

\bibitem{schrum2022mind}
Mariah~L Schrum, Erin Hedlund-Botti, Nina Moorman, and Matthew~C Gombolay.
\newblock Mind meld: Personalized meta-learning for robot-centric imitation learning.
\newblock In {\em 2022 17th ACM/IEEE International Conference on Human-Robot Interaction (HRI)}, pages 157--165. IEEE, 2022.

\bibitem{schuster2011research}
David Schuster, Scott Ososky, Florian Jentsch, Elizabeth Phillips, Christian Lebiere, and William~A Evans.
\newblock A research approach to shared mental models and situation assessment in future robot teams.
\newblock In {\em Proceedings of the Human Factors and Ergonomics Society Annual Meeting}, volume~55, pages 456--460. SAGE Publications Sage CA: Los Angeles, CA, 2011.

\bibitem{tabrez2020survey}
Aaquib Tabrez, Matthew~B Luebbers, and Bradley Hayes.
\newblock A survey of mental modeling techniques in human--robot teaming.
\newblock {\em Current Robotics Reports}, 1:259--267, 2020.

\bibitem{talone2015evaluation}
Andrew~B Talone, Elizabeth Phillips, Scott Ososky, and Florian Jentsch.
\newblock An evaluation of human mental models of tactical robot movement.
\newblock In {\em Proceedings of the Human Factors and Ergonomics Society Annual Meeting}, volume~59, pages 1558--1562. SAGE Publications Sage CA: Los Angeles, CA, 2015.

\bibitem{van2023tracking}
Basile Van~Hoorick, Pavel Tokmakov, Simon Stent, Jie Li, and Carl Vondrick.
\newblock Tracking through containers and occluders in the wild.
\newblock In {\em Proceedings of the IEEE/CVF Conference on Computer Vision and Pattern Recognition}, pages 13802--13812, 2023.

\end{thebibliography}

\end{document}